# Synthetic Error Dataset Generation Mimicking Bengali Writing Pattern

Md. Habibur Rahman Sifat, Chowdhury Rafeed Rahman, Mohammad Rafsan, and Md. Hasibur Rahman

United International University, Dhaka, Bangladesh
`msifat152028@bscse.uiu.ac.bd`, `rafeed@cse.uiu.ac.bd`,
`cargorafsan@gmail.com` `mrahman161260@bscse.uiu.ac.bd`

**Abstract.** While writing Bengali using English keyboard, users often make spelling mistakes. The accuracy of any Bengali spell checker or paragraph correction module largely depends on the kind of error dataset it is based on. Manual generation of such error dataset is a cumbersome process. In this research, We present an algorithm for automatic misspelled Bengali word generation from correct word through analyzing Bengali writing pattern using QWERTY layout English keyboard. As part of our analysis, we have formed a list of most commonly used Bengali words, phonetically similar replaceable clusters, frequently mispressed replaceable clusters, frequently mispressed insertion prone clusters and some rules for *Juktakkhar* (constant letter clusters) handling while generating errors.

## 1 Introduction

Bengali is the seventh most spoken language in the world [9]. There are approximately 228 million native speakers of Bengali [1]. Bengali language went under a thriving *Sanskritization* which started in the 12$^{th}$ century and continued throughout the middle ages. This resulted in the vast gap between the pronunciation and the script. In Bengali, there are also a large number of constant clusters or *Juktakkhars*. Owing to this complexity, there are two most common reasons for misspelling. One is the phonetic similarity of Bengali characters and another one is the difference between the representation of the grapheme and the phonetic utterances. For example, "সহজ" and "শহজ" both have the same pronunciation though the correct spelling is "সহজ". In our work, we consider "Avro" [2] for writing Bengali through English keyboard. Avro Keyboard supports phonetic layout named "Avro Phonetic" that allows us to type Bengali through the Romanized transliteration along with fixed keyboard layout. It is one of the most popular writing tools for Bengali. Here, we have selected "QWERTY" layout English keyboard, the most popular keyboard layout among users since early 1930 [8].

At present, there are a number of spell checkers available for Bengali such as - Soundex [11], Double Metaphone encoding [13], a hybrid of Soundex, Metaphone and string matching [10] and so on. The datasets on which these algorithms were tested were collected manually. Different users of Avro make mistakes in different ways. There is no study available on this issue. A robust analysis based automatic error word generation algorithm from correct word can come in handy while working with large amount of data. Introducing probability to such algorithm can help create varieties of error datasets from a large Bengali corpus within seconds, which in turn can help statistical data hungry machine learning based models to learn user behaviors for accomplishing various tasks.

In this research, we introduce a probability based algorithm for Bengali error dataset generation. Our algorithm mimics human writing error generating varieties of error with each run. In brief, our contributions are as follows:

1. Open source code for error word generation from input word
2. List of most commonly used *Juktakkhors* along with 17 rules for handling them for error introduction
3. List of phonetically similar replaceable clusters, frequently mispressed replaceable clusters and frequently mispressed insertion prone clusters

   All the above lists and the code are available for download [here](.).

## 2   RELATED WORK

A Bengali spell checker using Double Metaphone encoding was constructed in [13]. Another study [11] used Soundex code for error checking and fixing for Thai language. The Soundex algorithm was also used with Bengali language [12]. First, they converted Bengali word into Bengali phonetic code and then applied Soundex. Saha et al. [10] provided relevant suggestions for misspelled Bengali word correction. They combined edit distance, string matching, Soundex and Metaphone in their approach.

Mandal et al. proposed a Clustering-based approach for Bengali Spell checker [7]. They used Partitioning Around Medoids (PAM) algorithm [4]. Dynamic "Edit Distance Algorithm" was used in [3] for Kafi Noonoo language spell checking. The spell checker needs to consider different forms of the same word and can be time consuming. Kenneth et al. [6] developed a spell checker based on Shanon's noisy channel model which is used to detect misspelling in English.

All of these researches work on finding the misspelled words and error patterns [5] and try to correct them. For machine learning based context level spell checker development and for proper evaluation of any new word level spell checker, we work on error pattern analysis for Bengali writing and provide an algorithm for automatic error corpus generation. To the best of our knowledge, no such synthetic error generation algorithm has been proposed for Bengali language.

## 3   Our Dataset

We have collected 6.5 million sentences from various reputed news papers such as "Prothom Alo", "Noya Diganta" and "BDNews24" through web scraping from online. The publishing time ranges from 2017 to 2019. The corpus consists of various topics such as politics, sports, economics, entertainment and literature. We have selected 8637 most frequently occurring (appeared more than 1000 times in our corpus) words from our collected Bengali corpus for error dataset generation experiments. Some sample words along with two generated sample error words for each correct word have been provided in Section 5.

## 4 Methodology

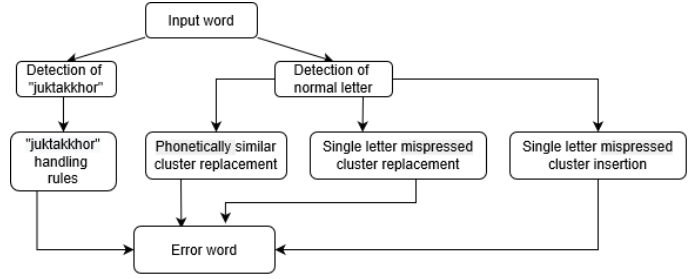

**Fig. 1.** Error Dataset Generation Algorithm Work Flow

The steps of error word generation from an input correct word in our algorithm have been shown in Figure 1. They are described as follows:

### 4.1 Distinguishing Between Normal Letter and *Juktakkhor*

We have selected total 170 *Juktakkhors* that are most commonly used in Bengali. We can identify a *Juktakkhor* if we check for the symbol ' ্ ' in between two alphabets. For example, regarding the word "পাঞ্জেরী -> প + ◌া + ে◌ + ন + ্ + জ + র + ◌ী", the normal letters are 'প', '◌া', 'ে◌', 'র', '◌ী' and the *Juktakkhor* is "ঞ্জ -> ন , ্ , জ".

We provide ten sample *Juktakkhors* in Table 1 with examples. The complete list has been provided here.

| sl. | column no. 1 |
|---|---|
| 1 | "ক্ট = ক + ্ + ট; যেমন- ডক্টর" |
| 2 | "ক্ত = ক + ্ + ত; যেমন- রক্ত" |
| 3 | "ক্র = ক + ্ + র; যেমন- চক্র" |
| 4 | "ক্ষ = ক + ্ + ষ; যেমন- পক্ষ" |
| 5 | "ক্ষ্ম = ক + ্ + ষ + ্ + ম; যেমন- লক্ষ্মী" |
| 6 | "ক্ল = ক + ্ + ল; যেমন- ক্লান্তি" |
| 7 | "ক্ক = ক + ্ + ক; যেমন- আক্কেল, টেক্কা" |
| 8 | "খ্র = খ + ্ + র; যেমন- খ্রিস্টান" |
| 9 | "ক্য = ক + ্ + য; যেমন- বাক্য" |
| 10 | "ক্ষ্ণ = ক + ্ + ষ + ্ + ণ; যেমন- তীক্ষ্ণ" |

**Table 1.** Some Words Containing *Juktakkhors*

The following three subsections deal with single letter errors.

### 4.2  Phonetically Similar Cluster Replacement

Phonetically similar alphabets in Bengali have the same pronunciation or phonetic utterance. If two of the letters have the same type of phonetic similarity, then these letters can be interchanged for making an error word. Soundex algorithm for word level Bengali spell checker proposed in [12] was based on this idea of phonetic similarity. Ten sample letters along with their replaceable letter siblings have been provided in Table 2. For example, if we want to write 'ক', we have to press letter 'k' which will suggest us 'ক'. Though, for the alphabet 'খ', we press 'kh' in Avro. The letter 'h' can be omitted by mistake and can turn into 'ক'. That is why, in row number 1, there is no replacement for 'ক' but for letter 'খ', the replacement is 'ক'. The full table has been given here.

| Serial No. | Alphabets | Replaceable Alphabets |
|---|---|---|
| 1 | 'ক' | [ ] |
| 2 | 'খ' | 'ক' |
| 3 | 'ঘ' | 'গ' |
| 4 | 'অ' | 'ও' |
| 5 | 'ঙ' | 'ং' |
| 6 | 'চ' | 'ছ' |
| 7 | 'ৎ' | 'ত' |
| 8 | 'থ' | 'ত', 'ট' |
| 9 | 'ঔ' | 'অউ' |
| 10 | 'স' | 'শ', 'ষ' |

**Table 2.** Phonetically Similar Cluster Replacement Table

The following two subsections deal with single letter errors that occur during fast typing.

### 4.3  Single Letter Mispressed Cluster Replacement

We are writing Bengali through English QWERTY keyboard. So, if we intend to write 'ক', we have to press button 'k' on the keyboard. As letter 'j' and 'l' are adjacent to letter 'k' in English QWERTY keyboard, 'জ' or 'ল' can accidentally replace 'ক'. Ten sample mispressed clusters are shown in Table 3. The full list has been given here.

### 4.4  Single Letter Mispressed Cluster Insertion

When we write a word of reasonably high length, there is a chance for a letter to be inserted by mistake. Suppose, a user wants to write "আমজনতা" using Avro. After pressing the letter 'j' for writing 'জ' the user can accidentally press letter 'k' and can insert 'ক' after 'জ' . This type of error is considered in the single letter mispressed cluster replacement class. As 'j' is at the right side of the keyboard, the immediate right key 'k' can be accidentally pressed. Similar analogy can be given for the keys situated to the left.

Bengali Error Dataset Generation 5| Serial No. | Alphabets | Replaceable Alphabets |
|---|---|---|
| 1 | 'ক' | ['ল', 'য'] |
| 2 | 'খ' | ['কগ', 'কজ', 'লহ','ঝ'] |
| 3 | 'গ' | ['ফ', 'হ'] |
| 4 | 'ন' | ['ব', 'ম'] |
| 5 | 'ল' | 'ক' |
| 6 | 'ধ' | ['দজ', 'দা', 'শ', 'ফহ'] |
| 7 | 'র' | ['এ', 'ে', 'ত'] |
| 8 | 'স' | ['আ', 'ো', 'দ'] |
| 9 | 'য' | ['ত', 'উ', 'ু'] |
| 10 | 'ষ' | ['সজ', 'সগ', 'আহ', 'ঢ'] |

**Table 3.** Single Letter Mispressed Cluster Replacement Table

| Serial No. | Alphabets | Insertion Prone Alphabet |
|---|---|---|
| 1 | 'ক' | 'ল' |
| 2 | 'খ' | 'গ' |
| 3 | 'গ' | ['ফ','হ'] |
| 4 | 'ড' | 'স' |
| 5 | 'ন' | 'ম' |
| 6 | 'ফ' | 'দ' |
| 7 | 'দ' | 'স' |
| 8 | 'ব' | 'ন' |
| 9 | 'শ' | 'জ' |
| 10 | 'ঈ' | 'অ' |

**Table 4.** Single Letter Mispressed Cluster Insertion Table

Ten such examples have been provided in Table 4. The full list of such probable mistakes have been provided here.

The following subsection deals with errors associated to typing Bengali constant clusters also known as *Juktakkhor*.

### 4.5 *Juktakkhor* Handling

Inspired from the spell checker based on Double Metaphone encoding in [13], we have analyzed Bengali *Juktakkhor* writing pattern from the perspective of *Juktakkhor* situation in a word and its pronunciation. We have come up with 17 rules regarding the generation of *Juktakkhor* related errors which are realistic. These 17 rules are as follows:

1. "জ্ঞ = জ + ঞ ", if this constant cluster 'জ্ঞ' is found in front of a word, it may change with letter 'গ' by mistake, for example, "জ্ঞান -> গান". Otherwise, no change is needed, for example "বিজ্ঞ".
2. "গ্য = গ + য", if it is found in the beginning,it could be "গা" by mistake. For example, "গ্যাস -> গাস". If found elsewhere, then it can convert to 'জ্ঞ'. Example: "ভাগ্য -> ভাজ্ঞ, ভাগ্যিস -> ভাজ্ঞিশ".

3. "চ্ছ = চ + ছ" can be mistakenly replaced by "চ্ছ -> ছছ/ ছ". Example: "লাচ্ছি -> লাছছি".
4. If there is 'য ফলা' -> '্য' in the beginning or at the middle of a word, it can be replaced with 'া' or 'ে'. For example, "ব্যবহার -> বাবহার, বেবহার". Also, if there is 'য ফলা' -> '্য' in the end of a constant cluster, it may not be in use. Example: "বাচ্য -> বাছছ or বাছ্য, ভাগ্য -> ভাগগ".
5. "স্ম" can be replaced with "স" or "শ" or with any of its cluster. Example: "স্মরণ -> শরন, সরণ".
6. "দ্ম" can be replaced with "দ্দ". For example, the word "পদ্ম" has "প + দ + ্ + ম" and here 'ম' is silent. So, this "পদ্ম" and "পদ্দ" have the same phonetic utterance.
7. If there are any letters along with "ম", those letters can be replaced with interchangeable clusters. Example: "সম্ভব -> সম্বব".
8. If there is "ব ফলা", then it can be neglected. Example: "তত্ত্ব -> তত্ত".
9. The letter "র" will be unchanged in case of "রেফ" or "র ফলা" but the letters along with "র" can be replaced with any interchangeable cluster letters. Example: "গ্রাম -> ঘ্রাম".
10. If "ক্ষ" is found in the beginning, it can be replaced only with "খ" but if any where else, then it can be replaced with "ক্ক" or "খ". Example: "ক্ষান্ত -> খান্ত, পক্ষ -> পক্ক".
11. In a constant cluster, if there is "ঙ" and after this there is a "া", then that constant cluster can be replaced with "ঙ্গ = ঙ + গ ". Example: "ব্যাঙাচি -> ব্যাঙ্গাচি".
12. If there is "ঙ" in a constant cluster, it can be replaced with "ং" and the letter along with it can be replaced with interchangeable cluster letters. Example: "ব্যাঙ -> ব্যাংগ".
13. In a constant cluster, if there is "ঞ" in the beginning, no changes needed, the letter along with it can be replaced with interchangeable cluster letters. Example: "লঞ্চ -> কঞ্চ" as "ল-> ক".
14. If the letter "হ" is added with letter "ন", then "হ" can be replaced with "ন" , in a constant cluster. For example, "চিহ্ন -> চিন্ন".
15. "ন্ন" can also be replaced with "হ্ন". Example: "চিন্ন -> চিহ্ন".
16. In a constant cluster, if there is same letter twice like "ল্ল, ক্ক", one of the letters can be neglected or both can be replaced with the interchangeable cluster. Example: "কল্লা -> কলা, অক্কা -> অকা".
17. If there are any letters along with "ল" / "ত" / "থ" / "দ" / "ধ" / "ট" / "ঠ" / "স" / "শ" / "ষ"/ "ন" / "ণ" , these letters can be replaced with interchangeable clusters. Example: "কষ্ট -> কস্ত" as "ষ্ট -> ষ + ্ + ট" , "বিষণ্ন -> বিশন্ন" and they do not have any replacements.

It is to note that the above rules have to be checked one by one in the given order for applicability. If any rule seems to be applicable, that particular rule is applied and all the rules that follow are ignored.

## 5  Results and Discussion

There are four probability values to tune in our algorithm. They are as follows:

– Phonetically similar cluster letter replacement probability (PP)
– Mispressed cluster letter replacement probability (MP)
– Juktakkhor change probability (JP)
– Mispressed cluster letter insertion probability (IP)

Through surveying on 60 Avro (a Bengali writing tool using English keyboard) users, we have fixed values of PP, MP, JP and IP as 0.25, 0.2, 0.3 and 0.2 respectively. The participants were satisfied with the generated synthetic errors on their typed correct sentences. Ten sample correct words along with two error words for each of them have been provided in Table 5.

| sl. | Correct word | Possible error words |
| --- | --- | --- |
| 1 | "কালিয়াকৈর" | "কালিয়াকইর , কালিয়াকৈড়" |
| 2 | "টাঙ্গাইল" | "তাংগাইল, তাঙ্গাইক" |
| 3 | "ট্রাকে" | "তেরাকে, ত্রাকে" |
| 4 | "পেট্রলবোমা" | "পেতরলবোমা, পেত্রল্বমা" |
| 5 | "তিনজন" | "তিণজন, তিনজোন" |
| 6 | "দগ্ধ" | "দগধ, দগদ" |
| 7 | "ব্যবসায়ী" | "বেবসায়ী, বেবসায়ি" |
| 8 | "ভর্তি" | "ভরতি, বর্তি" |
| 9 | "গ্রেপ্তার" | "গরেপ্তার, গেরেপতার" |
| 10 | "চেষ্টা" | "চেষতা, চেস্তা" |

**Table 5.** Algorithm Generated Sample Error Words

Insertion and mispressed cluster replacement take place mostly in words of length greater than three where each *Juktakkhor* is counted as only one letter. The number of these two kinds of errors generally do not occur more than once per word. We handle these cases in our algorithm. By tuning the values of PP, MP, JP and IP, one can make his dataset more or less error prone. For example, by making the value of IP equal to zero, one can exclude all sorts of insertion from his dataset.

Our error generation process is stochastic. As a result, with the same values of the four probability parameters and with the same corpus (a collection of words or sentences), multiple error datasets can be generated in seconds which will look completely different by simply running the algorithm multiple times.

## 6   Conclusion

In this research, we present a unique algorithm for generating Bengali error words from correct words which can be used for evaluating the performance of various word and context level spell checkers for Bengali language. The code has been made open source. As by product of the analysis, we have obtained important insights regarding *Juktakkhor* and single letter related errors which include replacement, insertion and deletion. We have also made these findings public through this research which will hopefully facilitate further research in Bengali. Future study can focus on the evaluation and comparison of existing Bengali spell checkers using our proposed spelling error generation algorithm on large Bengali corpus.

8     Md. Habibur Rahman Sifat